\documentclass[runningheads]{llncs}

\usepackage{eccv}

\usepackage{eccvabbrv}

\usepackage{graphicx}
\usepackage{booktabs}
\usepackage{wrapfig}
\usepackage[accsupp]{axessibility}  

\usepackage[pagebackref,breaklinks,colorlinks,citecolor=eccvblue]{hyperref}

\usepackage{orcidlink}
\usepackage{comment}

\renewcommand{\eqref}[1]{eq.~\ref{#1}}

\newcommand{\figref}[1]{Figure~\ref{#1}}

\newcommand{\secref}[1]{\S \ref{#1}}

\newcommand{\rebut}[1]{\textcolor{black}{#1}}

\begin{document}

\title{Self-supervised visual learning from interactions with objects} 


\author{Arthur Aubret\inst{1}\orcidlink{0000-0003-3495-4323} \and
Céline Teulière\inst{2}\orcidlink{0000-0002-7253-6221} \and
Jochen Triesch\inst{1}\orcidlink{0000-0001-8166-2441}}

\authorrunning{A.~Aubret et al.}


\institute{Frankfurt Institute for Advanced Studies \\ \email{lastname@fias.uni-frankfurt.de} \\ \and
Université Clermont Auvergne, Clermont Auvergne INP, CNRS, Institut Pascal, F-63000 Clermont-Ferrand, France\\
\email{celine.teuliere@uca.fr}}

\maketitle

\begin{abstract}
    Self-supervised learning (SSL) has revolutionized visual representation learning, but has not achieved the robustness of human vision. A reason for this could be that SSL does not leverage all the data available to humans during learning. When learning about an object, humans often purposefully turn or move around objects and research suggests that these interactions can substantially enhance their learning. Here we explore whether such object-related actions can boost SSL. For this, we extract the actions performed to change from one ego-centric view of an object to another in four video datasets. We then introduce a new loss function to learn visual and action embeddings by aligning the performed action with the representations of two images extracted from the same clip. This permits the performed actions to structure the latent visual representation. Our experiments show that our method consistently outperforms previous methods on downstream category recognition.
    In our analysis, we find that the observed improvement is associated with a better viewpoint-wise alignment of different objects from the same category.
    Overall, our work demonstrates that embodied interactions with objects can improve SSL of object categories.
  \keywords{Self-supervised learning \and Embodied learning \and Object representations}
\end{abstract}

\section{Introduction}

Self-supervised learning (SSL) methods acquire visual representations without access to labeled data. A central goal of SSL are representations that allow to discriminate different object categories. While much progress has been made \cite{chen2020simple}, most SSL methods lack the representation robustness of biological vision systems, e.g., with respect to the viewpoint of an object \cite{dong2022viewfool, purushwalkam2020demystifying}. This has spurred interest in time-based SSL methods that learn visual representations during extended interactions with objects. Such methods succeed in learning viewpoint-invariant object representations \cite{parthasarathy2023self,schneider2021contrastive,aubret2022time,yu2023mvimgnet}.


During learning about objects, humans frequently, e.g., rotate an object which they are holding. The issued motor commands determine how the visual input will change. Humans exploit this relationship between their motor commands and changes to the visual input to learn improved visual representations. For example, a study from \cite{harman1999active} considered two groups of subjects: individuals from a first ``active'' group watched and actively turned novel objects while members of a second ``passive'' group passively watched recordings of object interactions from the ``active'' group. When tested, the ``active'' group exhibited improved object recognition, suggesting that they had learned a more potent object representation. Since both groups watch the same sequence of views, this difference in object recognition may stem from the active group being aware of the applied action.

We hypothesize that the benefit of embodied manipulation for learning to classify objects is rooted in two facts. First, objects from the same class often have similar 3D shapes \cite{landau1988importance}. Similar manipulations applied to objects with similar 3D shapes produce similar changes to the visual input, which simply reflects the physics of the world. For example, consider a teapot seen from the front in an upright position, where one observes its snout. Turning the teapot by $180^\circ$ around the yaw axis would reveal its handle.
Most teapots share this property, such that understanding this relation between actions and changes to the visual input may support the emergence of a more semantic concept of ``teapot''. Importantly, the two views by themselves do not give access to this relation. Knowledge of the action connecting the two views is crucial.
Second, since natural interactions change the observer's perspective on a seen object, incorporating action-related information may increase the representation's sensitivity to an object's viewpoint compared to approaches aiming for viewpoint invariance. Such sensitivity to viewpoint, found in humans \cite{tarr1998three,vuilleumier2002multiple,hayward2003after}, may benefit category recognition. For instance, learning a viewpoint-invariant representation of a cup may dismiss features related to its handle, as a cup hides its handle when the handle is pointing away from the observer. Yet, this feature may be the only element that distinguishes a cup from a bowl, making it a key feature for categorizing a cup. 
\rebut{Here, we study the impact of actions changing objects' viewpoint on} visual representations learnt in an action-aware self-supervised fashion. We propose a new family of action-aware learning methods, Action-Aware Self-Supervised Learning (AA-SSL), which aligns a pair of image representations with a representation of the action applied to generate this pair. 
\rebut{We consider both object rotations and rotations around objects, i.e.\ ego-motion,} as parameterized actions and compare several state-of-the-art methods exploiting information about augmentations on large datasets that were previously under-used for learning visual representations. To better characterize the benefits of object interactions,  we also analyze the representations learnt on a new synthetic dataset of controlled $360^\circ$ object rotations, constructed with the close to photo-realistic simulation platform ThreeDWorld \cite{gan2021threedworld} and approximately 3,500 object models extracted from the Toys4k dataset \cite{stojanov2021using}. 

Our experiments demonstrate that AA-SSL consistently outperforms several state-of-the-art methods on visual representation learning with object interactions. Our analysis shows that AA-SSL finds a better trade-off between viewpoint invariance/sensitivity than previous approaches and better aligns different objects within a category based on their orientation. We also find that AA-SSL is more robust to the absence of augmentations that are usually considered critical in visual SSL. \rebut{In summary, we:}
\begin{enumerate}
    \item propose to simultaneously learn image embeddings and action representations while manipulating or moving around an object;
    \item \rebut{demonstrate the superiority of this approach over state-of-the-art baselines};
    \item \rebut{provide an in depth-analysis of these results.}
\end{enumerate}

 Overall, our work highlights the benefits of action-aware learning for building more robust and semantic visual representations. We will release the source code in \href{https://github.com/trieschlab/AASSL}{https://github.com/trieschlab/AASSL}.
\section{Related works}

\paragraph{Data-augmented SSL.} Many state-of-the-art SSL methods strive to learn image representations that are invariant to data-augmentations like cropping/resizing an image or adding color jitter to an image \cite{chen2020simple,he2020momentum,grill2020bootstrap,bardes2022vicreg}. More specifically, they learn image embeddings for which an image and its augmented variant, also called positive pair, are close, while both being overall distant from all other image embeddings. Following the same idea, multi-modal SSL methods try to align the representation of one sensory modality, e.g., images, with those of another modality, e.g., audio \cite{mittal2022learning,morgado2021audio}, touch \cite{zambelli2021learning} or text \cite{radford2021learning}. Furthermore, SSL methods can be combined with heterogeneous pretext tasks \cite{doersch2017multi} or reconstruction-based SSL methods \cite{hernandez2023visual}. However, these approaches do not leverage the ability to interact with objects to learn object representations.

\paragraph{SSL through time.} Rooted in biological models of learning invariant representations \cite{franzius2011invariant,wiskott2002slow}, there has recently been a surge of interest in approaches that learn invariant visual representations by exploiting temporal coherence. These approaches learn image embeddings that are (almost) invariant through time during natural interactions with objects and show great benefits for downstream action recognition \cite{knights2021temporally}, category recognition \cite{aubret2022time,parthasarathy2022self,orhan2020self,sanyal2023computational,yu2023mvimgnet,purushwalkam2020demystifying}, object instance recognition \cite{schneider2021contrastive,purushwalkam2020demystifying}, scene understanding \cite{parthasarathy2023self,gordon2020watching} and object tracking/segmentation \cite{xu2021rethinking,gordon2020watching}. The representations also demonstrate strong robustness to deformations and transfer well to categorization tasks in other datasets \cite{parthasarathy2023self,tschannen2020self}. On top of SSLTT, \cite{wu2021contrastive} propose to exploit the frame similarity across different videos to generate stronger positive pairs and \cite{jayaraman2016slow} introduce a higher order temporal coherence by making feature changes between frames stable. In practice, this line of work can learn view-invariant representations, such that clusters of individual objects can appear in the representation \cite{schneider2021contrastive}. We show in 
\secref{sec:experiments} that this strategy is sub-optimal for category recognition, as keeping features specific to an object's orientation may convey important information about its category membership.


\paragraph{Exploiting knowledge of augmentations in SSL.} Previous approaches have explored the idea of retaining information about the transformation applied on an image to improve SSL, but these studies were restricted to conventional image transformations. A first line of works learns to predict an applied augmentation, from a single image \cite{dangovski2021equivariant} or two augmented images \rebut{using an inverse model} \cite{ciperxu,lee2021improving}. A second line of works learns embeddings that are equivariant to data augmentations using a predictor that learns to predict the next embedding of an image after applying an augmentation \cite{devillers2022equimod,garrido2023self,bhardwaj2022steerable,przewikezlikowski2023augmentation}. These ideas were also applied for video representation learning \cite{jenni2021time, wang2016actions}. Here, we extract egocentric actions during natural interactions with objects and show in \secref{sec:experiments} the superiority of our method when adapting previous methods to work with natural actions. Another work leverages the possibility to apply twice the exact same augmentation on two images to structure the visual representation \cite{gupta2023learning}, which is impossible in our case. Further works empirically show that the kind of data-augmentations to which a representation should be invariant usually depends on the downstream task \cite{xiao2020noise} and that keeping subparts of the representation that vary with respect to the applied augmentations tends to improve the visual representation \cite{xie2022should,wang2023distortion}.

\paragraph{Embodied actions for visual representation learning}

\rebut{A body of work implements inverse models to achieve robotic \cite{agrawal2016learning} or reinforcement learning tasks \cite{pathak2017curiosity,kim2019emi,efroni2021provably}. These works do not try to learn object representations. Two kind of approaches notably leverage egocentric natural actions for the purpose of object categorization. The former considers egomotion as an action and learns visual features with an inverse model \cite{agrawal2015learning,liang2023alp}. The second one learns equivariant representations from either a moving car or rotating objects \cite{jayaraman2015learning,jayaraman2016look,jayaraman2017learning}. However, their experiments on objects seen through different viewpoints include a limited set of objects, which do not allow to properly evaluate downstream category recognition. We further find that AA-SSL outperforms raw action prediction and action equivariance in \secref{sec:experiments}}. A very recent work learns representations of actions defined as the inertial head motion \cite{tan2024egodistill} to fuse action and visual representations and learn video-clip representations that support high-level {\em action recognition}. Our work focuses on learning strong object representations by incorporating action information. In addition, they use a frozen (ImageNet-pretrained) visual encoder and only train the action encoder. In contrast to them, we simultaneously learn image and action encodings.







\begin{figure*}
    \centering
    \includegraphics[width=1\linewidth]{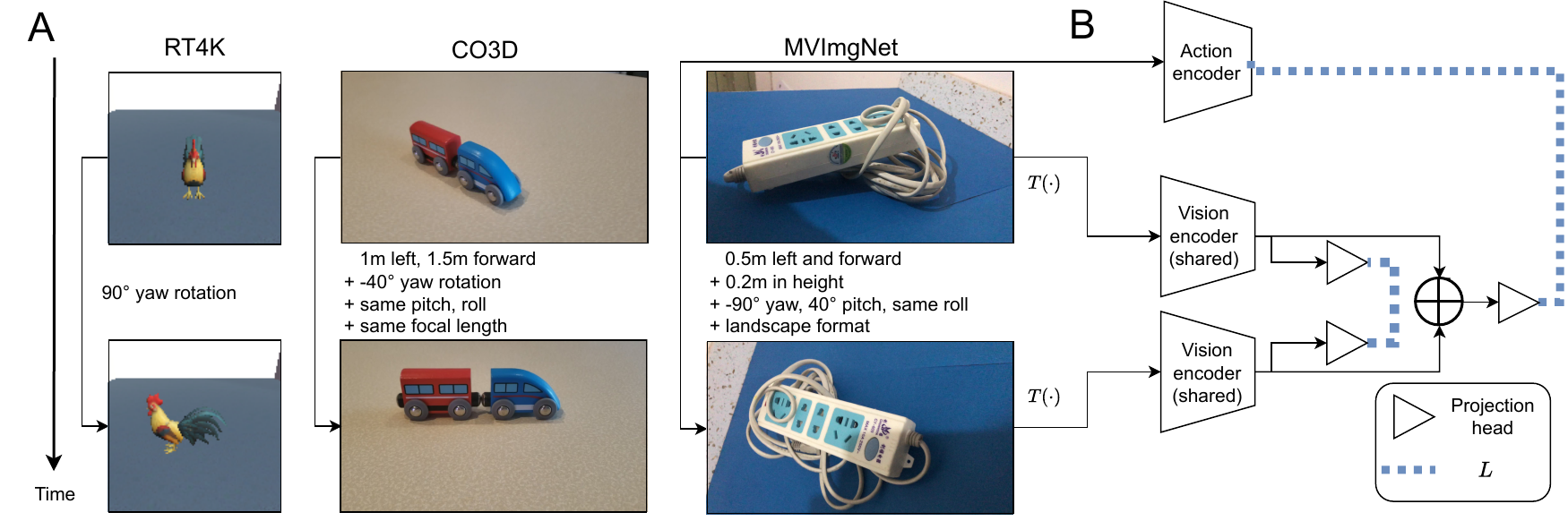}
    \caption{A) Example of object interactions from four datasets. B) Summary of the learning process of AA-SSL. See text for details.}
    \label{fig:learnarc}
\end{figure*}

\section{Methods}
We study how \rebut{generating different views of objects by} interacting with them, rather than passively watching them, impacts visual representations and category generalization. We introduce/modify three datasets showing extended interactions with objects, which incorporate information about changes of camera pose or objects' orientations (Fig.~\ref{fig:learnarc}A)  This allows us to extract/estimate the 3D action applied to go from one image to another. Then, we introduce our new Action-Aware Self-Supervised Learning (AA-SSL) method (Fig.~\ref{fig:learnarc}B). AA-SSL simultaneously learns action and image embeddings such that action embeddings become aligned with the concatenation of the image representations that precede and follow an action, using SOTA augmentation-based SSL methods. \rebut{In contrast to previously proposed inverse models \cite{agrawal2015learning,ciperxu}, AA-SSL learns to predict a learnt representation of actions.}


\subsection{Video datasets of interactions with objects}\label{sec:datasets}

In this section, we describe our datasets and how we extract the action $a_{t, t'}$ that leads to visual changes between two temporally related images $x_t$ and $x_{t'}$. In our datasets, this action is either an object rotation or a movement around an object, both shown in a video clip. 

\paragraph{Revolve Toys4k (RT4K) dataset.} This dataset contains first-person views of the 3574 objects extracted from the Toy4k dataset \cite{stojanov2021using}. To build this dataset, we simulate manipulations of objects seen from slightly above; we record for each of them its view and orientation every two degrees of rotations during a yaw-axis revolution, resulting in 180 images per object. We split in an object-wise fashion the dataset into 375,300 train images and 268,020 test images. To extract the 3D action $a_{t, t'}$ applied between two views of an object, we extract the rotation angle induced by object's orientations and compute the sine and cosine of the angle.

\paragraph{CO3D dataset.} While RT4K contains manipulations of objects, the CO3D dataset simulates the egocentric motion of an observer around an object \cite{reizenstein2021common}. It includes 1.5 million frames distributed across 19,000 videos and 51 categories of objects. We keep one out of every two images and apply a $75\%/25\%$ train/test object split. To encode the applied action $a_{t, t'}$ between two images of the same video, we compute the quaternion of the rotation matrix between two views and calculate the translation vector between the views.



\paragraph{MVImgNet dataset.} Similarly to the CO3D dataset, MVImgNet shows objects during a turning movement of the camera around objects \cite{yu2023mvimgnet}, It contains approximately 6.5 million frames from to 219,188 videos of 238 real-life categories. Thus, it is one order of magnitude bigger than CO3D. We keep 80\% of videos for training and 10\% each for validation and test sets. Similarly to the CO3D dataset, we compute the quaternion of the rotation matrix between two views, calculate the translation vector between the views and also add a binary number, constant through a video, indicating whether the camera is held vertically or horizontally. These three component form $a_{t, t'}$.

We also consider two smaller versions of MVImgNet. On a small (MVImgNet-S) and tiny version (MVImgNet-T) of MVImgNet, we train on only $10\%$ and $2\%$ of objects, respectively. We denote the full version as MVImgNet-F and the three versions as MVImgNet.

\subsection{Action-Aware Self Supervised Learning}

We aim to test the hypothesis that exploiting information available during interactions with objects supports visual representations with better category generalization. To leverage action information, we propose a new SSL architecture, action-aware SSL (AA-SSL), that easily plugs onto previous augmentation-based SSL methods (Fig.~\ref{fig:learnarc}).

Each training batch $\mathcal{B}$ contains triplets $(x_t, x_{t'}, a_{t,t'})$, where $x_t$ is a sampled image, $x_{t'}$ is another image randomly sampled in the same video clip as $x_t$ and $a_{t, t'}$ is the action applied between two temporally related images (cf. \secref{sec:datasets} for details about how we construct $a_{t, t'}$). It results that $x_t$ and $x_{t'}$ show the same object.

%
%
Let $f_{\theta}$ be our visual backbone encoder, for all tuples $(x_t, x_{t'}, a_{t,t'})$, we compute their latent visual representations $h_t=f_{\theta}(T(x_t))$ and $h_{t'} = f_{\theta}(T(x_{t'}))$ where $T$ is a standard image transformation sequence used in SSL, e.g., image crop, image resize and color distortion. These representations will serve to compute two loss functions. Using a projection head $g$, we apply an arbitrary contrastive SSL loss function $L(g(h_t),g(h_{t'}))$ \cite{aubret2022time} (e.g.\ SimCLR \cite{chen2020simple} or VICReg \cite{bardes2022vicreg}).

Our contribution resides in the introduction of a second complementary term that learns to align two action embeddings computed in different ways. First, we compute an action embedding from the raw action encoding as $g^1_{\phi^1}(a_{t,t'})$ where $g^1_{\phi^1}$ is the action encoder. Second, we compute another action embedding from the concatenation of the two temporally related views' representations $g^2_{\phi^2}(h_t,h_{t'})$, where $g^2_{\phi^2}$ is an action projection head. Thereafter, we can minimize $L(g^1_{\phi^1}(a_{t,t'}), g^2_{\phi^2}(h_t,h_{t'}))$ to train the weights of the action encoder, action projection head and visual encoder. For a given pair, the whole loss function reads:
\begin{equation}
     \mathcal{L}_{\rm AA}(x_t, x_{t'}) = L(g_{\phi}(h_t),g_{\phi}(h_{t'})) + \lambda L(g^1_{\phi^1}(a_{t,t'}), g^2_{\phi^2}(h_t,h_{t'})),
\end{equation}
where $\lambda$ is a weighting factor ($ \lambda = 1$ in the following). This way the visual backbone is trained with both the invariance and action-aware losses.

\subsection{Hyper-parameters and evaluation}\label{sec:hyperparameters}
For each dataset, we optimize the minimal crop size in $\{0.07, 0.2, 0.5\}$ (on SimCLR and VICReg) \cite{parthasarathy2022self,aubret2022time} and find $0.2$ to be the best value in all datasets. We also use different hyper-parameters for the invariance-based and action-aware SSL loss functions ($\tau_i$ and $\tau_a$ in SimCLR and $\lambda_i,\, \mu_i,\, \nu_i,\,\lambda_a,\, \mu_a,\, \nu_a$  in VICReg). We selected the best SimCLR temperature hyper-parameters in $\{0.07, 0.1, 0.5\}$ and the best VICReg hyper-parameters in ${(1,1), (10,10), (25,25)}$ with $\nu=1$ kept fixed. We found $\tau_i=0.1$ and $(\lambda_i, \mu_i)=(25,25)$ to be the best parameters in all datasets. Unless stated otherwise, other hyper-parameters were not varied \rebut{and were tuned with SimCLR-TT}, thereby putting our method at a disadvantage.


For RT4K, the backbone encoder $f_{\theta}$ is a ResNet18 and projection heads/action encoders are constructed following (Linear(256), BatchNormalization, ReLU, Linear(128)). We train for 40 epochs and optimize the loss with AdamW, a learning rate of $5e-4$, a batch size of $2 \times 128$, set $\tau_a=0.1$ and $(\lambda_a, \mu_a)=(25,25)$.

For CO3D and MVImgNet we encode the visual representation with a ResNet50 and use projection heads/action encoders with 2 hidden layers of 1024 neurons. We use one hidden layer for MVImgNet-F. We train for 200 epochs, 200 epochs, 100 epochs and 50 epochs in respectively CO3D, MVImgNet-T, MVImgNet-S and MVImgNet-F with AdamW optimizer, a cosine learning rate decay starting at $0.001$, a batch size of $2\times 256$ in CO3D, a batch size of $2\times 512$ in MVImgNet, $\tau_a=0.5$, $(\lambda_a, \mu_a)=(10,10)$. 


To evaluate the learned representations, \rebut{we follow standard evaluation protocols }\cite{chen2020simple,bardes2022vicreg}. We freeze the backbone encoder  $f_\theta$ (unless stated otherwise) and train a linear classifier on top of its output representations \rebut{to predict category labels}. We train the linear classifier on the train split and evaluate it on the test split. 

\section{Results}\label{sec:experiments}

We first compare AA-SSL to previous SSL methods for downstream categorization. Then we study the role of viewpoint sensitivity/invariance of the different methods as a potential explanation of the results. Finally, we qualitatively and quantitatively analyze the generalization properties of AA-SSL and its robustness to spurious correlations in positive pairs. 

\begin{table*}[]
    \centering
    \caption{Test \rebut{category} accuracy \rebut{(linear classifier)} with representations learnt through different self-supervised methods on different datasets. Due to computational constraints, we run 3 seeds only on RT4K, MVImgNet-T and MVImgNet-S. On MVImgNet, each seed uses a different subset of train objects, shared across methods. Preliminary analysis showed close-to-zero standard deviation in CO3D after convergence for all methods ($\pm 0.2$).}
    \begin{tabular*}{\linewidth}{@{\extracolsep{\fill}}ccccccccc}
        \toprule
          & SimCLR & -TT(direct) & -TT & \textbf{AA-} & -CIPER & -EquiMod \\ \midrule
         RT4K & $71.1 \pm 0.5$& $72.3 \pm 0.3$& $70.0 \pm 0.7$& $\mathbf{74.8 \pm 0.3}$ & $71.3 \pm 1.3$& $69.7 \pm 0.4$ \\
         CO3D & $79.6$& $79.6$& $80.2$& $\mathbf{81.3}$& $80.8$& $78.9$ \\
        MVImgNet-T & $40.1 \pm 0.6$& $39.6 \pm 0.3$& $39 \pm 0.7$& $\mathbf{42.8 \pm 0.8}$& $40.1 \pm 1.4$& $36 \pm 0.2$ \\
         MVImgNet-S & $73.2 \pm 0.1$& $73 \pm 0.2$& $77.3 \pm 0.3$& $\mathbf{78.4 \pm 0.1}$ & $76.8 \pm 0.4$& $75.4 \pm 0.2$ \\
         MVImgNet-F & $93.4$ & $94.2$ & $\mathbf{96.4}$& $96.1$& $96.2$& $96.3$ \\
          \midrule
          & VICReg & -TT(direct) & -TT & \textbf{AA-} & -CIPER & -EquiMod \\ \midrule
         RT4K & $69.8 \pm 0.8$& $69.5 \pm 0.3$& $67.2 \pm 0.9$& $\mathbf{70.8 \pm 0.4}$& $68.5 \pm 0.8$& $66.9 \pm 0.1$ \\
         CO3D& $78.2$& $76.6$& $77.2$& $\mathbf{80.6}$& $79.8$& $76.7$ \\ \bottomrule
    \end{tabular*}
    \label{tab:mainresults}
\end{table*}

\subsection{AA-SSL outperforms standard SSL}\label{sec:groundres}

To evaluate the visual representations learnt by AA-SSL (AA-*), we compare it to several SSL baselines that do not account for neither time, nor action (SimCLR and VICReg) and previously introduced variants that also augment an image by sampling a positive pair among the direct predecessor/successor (*TT(direct)) or uniformly in the same clip (*-TT) \cite{aubret2022time,schneider2021contrastive}. We also select two representatives of the related work that exploit data-augmentations parameters during learning. We adapt them to use parameterized actions. The first \rebut{is an inverse model that} predicts the normalized raw action (*-CIPER) \cite{ciperxu} and the second \rebut{is an equivariant model that predicts} the next representation following an action (*-EquiMod) \cite{devillers2022equimod}. We apply crop/resize, grayscaling and color jittering in all methods with same parameters as in \cite{chen2020simple}. We apply horizontal flipping with SimCLR/VICReg only in CO3D and MVImgNet, assuming using ``more'' augmentations should not drastically impact the representation when learning on large datasets. 

In \tableautorefname~\ref{tab:mainresults}, we observe that AA-SSL consistently learns better representations than previous approaches when combined with both VICReg and SimCLR. We note two exceptions. In MVImgNet-F, the huge number of objects elicit a strong performance in baselines, making hard to discover benefits.

We further verify the compatibility of AA-SSL with the non-contrastive BYOL method \cite{grill2020bootstrap} on RT4K and find that AA-BYOL obtains 71.3\% accuracy versus 69\% for BYOL and 67.8\% for BYOL-TT. Interestingly, unlike previously reported results \cite{aubret2022time}, SimCLR and VICReg achieve very good performance even without an implicit form of temporal invariance learning. Overall, we conclude that AA-SSL during object interactions generally boosts self-supervised visual representation learning.

\begin{table*}
\caption{Transfer learning accuracy with models pre-trained on MVImgNet-F with SimCLR. For ImageNet experiments, we finetuned only a linear classifier for 50 epochs. We include SimCLR and the two strongest comparison baselines.}
\label{tab:transfer}
\begin{tabular*}{\columnwidth}{@{\extracolsep{\stretch{1}}}*{7}{c}@{}}
\toprule
                               Dataset & SimCLR & -TT    & -CIPER & \textbf{AA-} \\ \midrule
\multicolumn{1}{c}{ImageNet-1K}     & $40.3$ & $42.4$ &  $43.5$  & $\mathbf{44.3}$          \\ 
\multicolumn{1}{c}{ImageNet-100}    & $61.8$ & $64.4$ & $65.2$  &  $\mathbf{66}$          \\
\multicolumn{1}{c}{Tiny-ImageNet}   & $35.1$ & $36.6$ & $\mathbf{38.6}$           & $38$   \\ 
\bottomrule
\end{tabular*}
\end{table*}

\subsection{AA-SSL improves transfer learning}

In \tableautorefname~\ref{tab:transfer}, we display transfer learning results with models pre-trained on MVImgNet-F. We observe that AA-SimCLR outperforms other baselines on reference datasets for category recognition (ImageNet-1K, ImageNet-100). It performs close to SimCLR-CIPER on Tiny-ImageNet. Overall, it confirms that AA-SimCLR learns better visual representations. 

\subsection{AA-SSL benefits from time invariance }
\vspace{-0.5cm}
\begin{table*}[]
    \centering
    \caption{Test \rebut{category} accuracy \rebut{(linear classifier)} trained with one frozen view representation per video.}
    \begin{tabular*}{\linewidth}{@{\extracolsep{\fill}}ccccccccc}
        \toprule
          Dataset & SimCLR & -TT(direct) & -TT & \textbf{AA-} & -CIPER & -EquiMod \\
          \midrule
         RT4K & $63.9 \pm 0.2$& $65.0 \pm 0.3$& $67.8 \pm 0.8$& $67.8 \pm 0.1$& $\mathbf{68.6 \pm 0.7}$& $67.7 \pm 0.4$ \\
         CO3D & $74.8$ & $75.2$& $76.2$& $\mathbf{77.4}$ & $76.7$& $73.7$ \\
        MVImgNet-T & $33.5 \pm 0.7$& $32.7 \pm 0.4$& $35 \pm 0.5$& $\mathbf{36 \pm 0.5}$& $34.5 \pm 1.9$& $32.7 \pm 0.6$ \\
        MVImgNet-S & $65.5 \pm 0.4$& $64.5 \pm 0.6$& $\mathbf{73.6 \pm 0.1}$& $70.2 \pm 2.1$& $71.4 \pm 0.8$& $70.3 \pm 1.3$ \\ 
         MVImgNet-F & $91.6$& $92.6$& $\mathbf{96.2}$& $95.7$& $95.3$& $96$ \\
        \midrule
          Dataset & VICReg & -TT(direct) & -TT & \textbf{AA-} & -CIPER & -EquiMod \\
          \midrule
         RT4K & $63.5 \pm 1.1$ & $64.7 \pm 0.4$& $63.7 \pm 0.6$& $\mathbf{65.4 \pm 0.1}$& $64.7 \pm 0.6$& $63.0 \pm 0.5$ \\
         CO3D & $71.3$ & $70.6$& $71.1$& $\mathbf{76.3}$& $75.6$& $70.3$ \\
         \bottomrule
    \end{tabular*}

    \label{tab:mainresultsS}
\end{table*}

To study whether AA-SSL can still benefit from time invariance, we assess the representation by learning a linear classifier with only one image per training video (and thus object). For this linear finetuning stage, we reuse the models learnt in \secref{sec:groundres}. We hypothesize that such setting will make it harder to the linear classifier to learn viewpoint invariance. In \tableautorefname~\ref{tab:mainresultsS}, SSLTT (*-TT) demonstrates categorization at least on par with VICReg and SimCLR, presumably because the linear classifier better benefits from viewpoint invariance in the representation if it can not learn it by itself. Importantly, AA-SSL performs at least similarly with time-invariant methods and outperforms SimCLR and VICReg, suggesting that it still takes the best out of sheer invariance through time.

\subsection{\rebut{The advantage of AA-SSL remains with different architectures of projection heads}}

\rebut{We want to verify how results observed in \secref{sec:groundres} scale with the capacity of projection heads. In \tableautorefname~\ref{tab:heads}, we select the 2 best-performing comparison baselines on MVImgNet-S, namely SimCLR-TT and SimCLR-CIPER and compare them with AA-SimCLR for different architectures of projection heads. We clearly observe that AA-SimCLR outperforms SimCLR-TT and SimCLR-CIPER for all projection heads. We also observe that all methods have best performance with two layers in both projection heads, but AA-SimCLR is much more robust to smaller projection heads.}

\rebut{Previous works showed that the optimal layer for downstream categorization can change according to the depth of the projection head \cite{chen2020big}. In \tableautorefname~\ref{tab:layers}, we take the models trained with $3$ hidden layers of $1024$ neurons for both projection heads to investigate the potential of hidden layers for downstream categorization.} In line with previous work \cite{chen2020big}, we find the output of the backbone encoder is the best layer to do linear fine-tuning on categories. In addition, AA-SimCLR outperforms SimCLR-TT and SimCLR-CIPER across all layers. \rebut{Overall, we conclude that the performance advantage of AA-SSL does not depend on the architecture of the projection head or the network layer used for evaluation.}

\begin{table}[]
\caption{\rebut{MVImgNet-S test category accuracy for SimCLR-TT, AA-SimCLR, SimCLR-CIPER, when trained with different architectures of projection heads. We select the best test accuracy, computed every 5 epochs during downstream linear fine-tuning. On the x-axis, we describe projection heads as $\texttt{Number hidden neurons} \times \texttt{Number hidden layers}$. The first and second rows refer to the projection heads of the invariant and action-aware losses, respectively. Hence, for SimCLR-TT, only a change in the first row entails a different training setting.}}
    \centering
    \begin{tabular*}{\linewidth}{@{\extracolsep{\fill}}ccccccc}
        \toprule
        & $1\times256$ & $1\times1024$ & $1\times1024$ & $1\times1024$ & $2\times1024$ & $3\times1024$ \\
                & $1\times256$ & $1\times1024$ & $2\times1024$ & $3\times1024$ & $2\times1024$ & $3\times1024$ \\
            \midrule
         SimCLR-TT & $68.6$& $73.1$& $73.1$& $73.1$& $77.3$& $77.1$ \\
         AA-SimCLR & $\mathbf{73.4}$& $\mathbf{77.1}$& $\mathbf{77.2}$& $\mathbf{77.6}$& $\mathbf{78.3}$& $\mathbf{77.7}$ \\
         SimCLR-CIPER & $71.7$& $73.6$& $75.4$& $75.9$& $76.8$& $76.5$ \\
         \bottomrule
    \end{tabular*}
    \label{tab:heads}
\end{table}

\vspace{-0.8cm}

\begin{table}[]
    \caption{\rebut{Test category accuracy of a linear classifier probed on hidden layers of projection heads after ReLU activation. We select the best test accuracy, computed every 5 epochs during downstream linear fine-tuning. ``Inv-n" and ``AA-n" refer to the n-th ReLU activation of the invariant and action-aware heads, respectively. Since the action-aware head takes in two images representations, we duplicate a given image representation.}}
    \begin{tabular*}{\linewidth}{@{\extracolsep{\fill}}cccccccc}
    \toprule
    Method &
      Rep &Inv-1 &Inv-2 &Inv-3 &AA-1 &AA-2 &AA-3 \\ 
    \midrule
    SimCLR-TT &$77.1$ &$73.7$ &$69.2$ &$63.2$ &NA &NA &NA \\ 
    \textbf{AA-SimCLR} &$\underline{\mathbf{77.7}}$ &$\mathbf{74.7}$ &$\mathbf{70.7}$ &$\mathbf{64.7}$ &$\mathbf{72}$ &$\mathbf{64.1}$ &$\mathbf{38}$ \\ 
    SimCLR-CIPER &$76.5$ &$73.3$ &$69.1$ &$63.1$ &$70$ &$61.9$ &$18.9$ \\ 
      \bottomrule
    \end{tabular*}
    \label{tab:layers}
\end{table}

\subsection{Viewpoint sensitivity partly explains improved accuracy of AA-SSL}\label{sec:invariance}


\begin{figure*}
    \centering
    \includegraphics[width=1\linewidth]{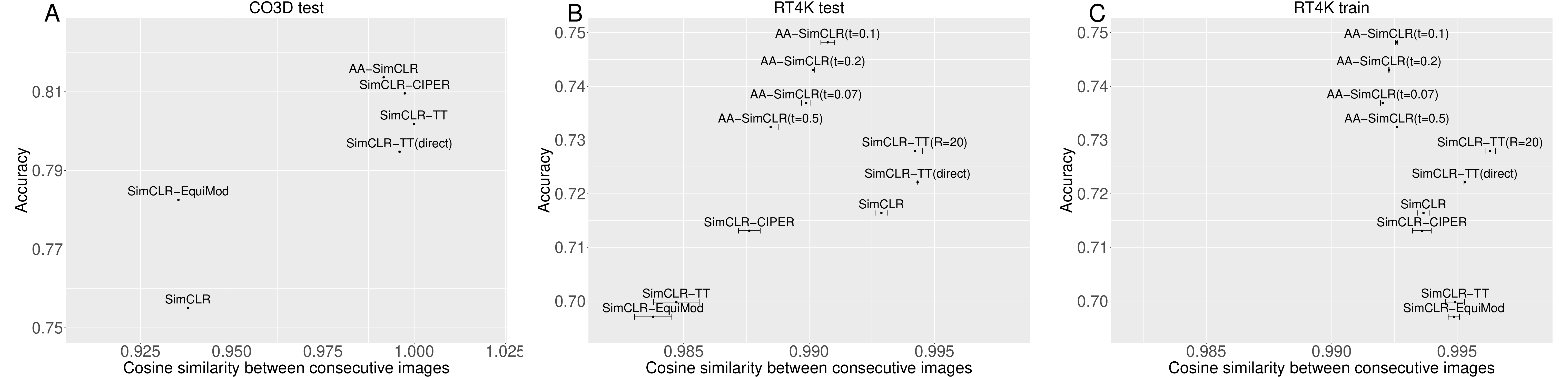}
    \caption{Scatter plot of categorization accuracy vs.\ level of invariance in the representation (measured as the cosine similarity of the representations of adjacent video frames) for several methods based on SimCLR. We compute the level of invariance on A) the test set of CO3D, B) the test set of RT4K and C) the train set of RT4K. $t$ indicates the SimCLR temperature applied for learning the action representation and $R$ denotes the maximal rotation angle between an image and its positive pair during training. Horizontal bars show the standard deviation over the 3 seeds, when available. The best performing AA-SSL shows an intermediate level of invariance.}
    \label{fig:accuracydistance}
\end{figure*}
We hypothesize that the improved generalization over categories of AA-SSL may originate from its level of viewpoint sensitivity in the representation. We verify this in \figureautorefname~\ref{fig:accuracydistance}A-B) , where we plot the accuracy of different methods against their average cosine similarity between the representations of two consecutive video frames in the test set. We observe that the peak of accuracy corresponding to AA-SimCLR always situate between strong invariance and important view sensitivity. Methods or parameter settings leading to either very low and very high viewpoint invariance respectively show very poor and poor performance. This suggests that AA-SimCLR finds a better viewpoint invariance/sensitivity trade-off than previous methods. Interestingly, in RT4K (\figureautorefname~\ref{fig:accuracydistance}B), AA-SimCLR becomes more viewpoint sensitive than SimCLR, despite explicitly optimizing a loss pushing for invariance. We hypothesize that the large number of symmetrical objects present in the Toys4k model library incites AA-SimCLR to discover view-specific features. 

In Figure \ref{fig:accuracydistance}B-C), we see that the level of invariance of the representation can drastically vary between the training and test sets in RT4K, in particular for SimCLR-TT and SimCLR-EquiMod. This indicates that methods with too strong incentives for invariance through time may be prone to over-fitting when training on relatively small datasets, which hinders generalization over categories.

Despite these correlations, we observe that the SimCLR temperature for AA-SimCLR largely impacts these gaps, even for a given level of invariance. Overall, we conclude that while AA-SSL helps avoiding over-fitting and finds a better trade-off between viewpoint invariance/sensitivity, this does not fully explain the performance advantage of AA-SSL.

\subsection{AA-SSL aligns representations of similar views of different objects from the same category}\label{sec:quality}

\begin{figure*}
    \centering
    \includegraphics[width=1\linewidth]{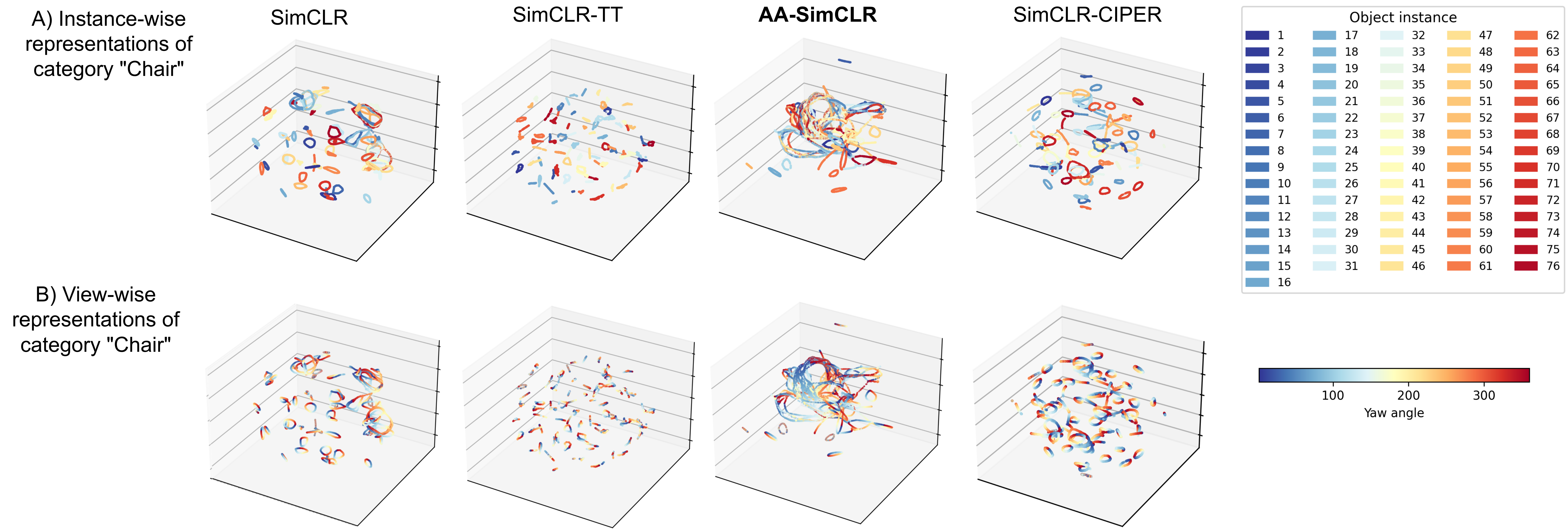}
    \caption{PaCMAP \protect\cite{wang2021understanding} visualization of all embeddings of test images in category ``Chair'' in RT4K. Representations are colored according to A) the object instance associated to their image and B) the object yaw orientation. Note how only AA-SimCLR aligns representations of similar views of different objects from the same category.}
    \label{fig:pacmap}
\end{figure*}

To further explore the properties of representations learnt with AA-SSL, \figureautorefname~\ref{fig:pacmap} shows PaCMAP projections of the representations of objects from the \textit{chair} category (which contains the highest number of objects in RT4K). 
While SimCLR-TT displays dense clusters of instances' images, AA-SimCLR, SimCLR and SimCLR-CIPER learn representations that capture the proximity, in terms of angle, between viewpoints. In \figureautorefname~\ref{fig:pacmap}B), we find that AA-SimCLR is the only method that aligns viewpoints of different object instances with each other, such that two chairs observed from the same viewpoint are perceived as more similar than one chair seen from two significantly different angles. Since SimCLR-CIPER does not exhibit this effect, this suggests that learning an action representation may be crucial for this kind of generalization and the resulting improvements in category recognition.

\begin{figure*}
    \centering
    \includegraphics[height=2.5cm]{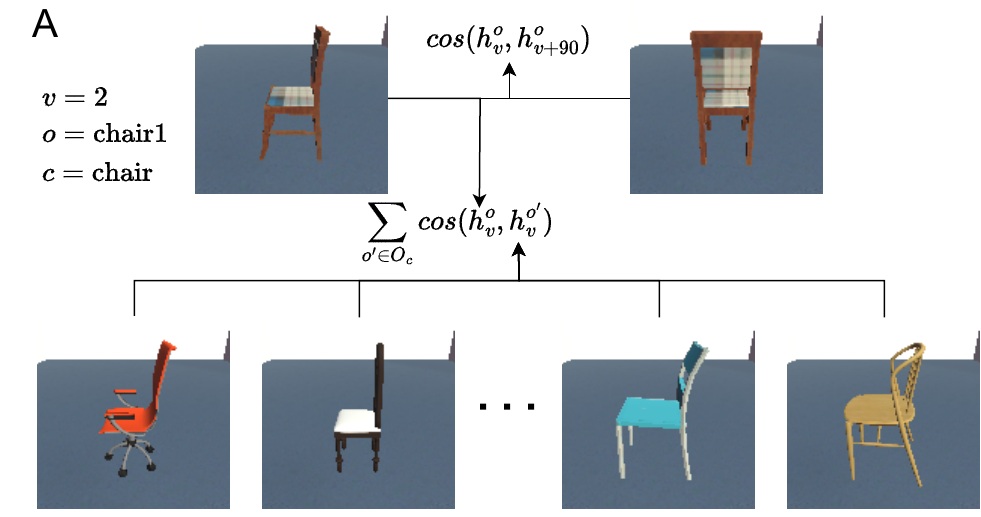}
    \includegraphics[height=2.5cm]{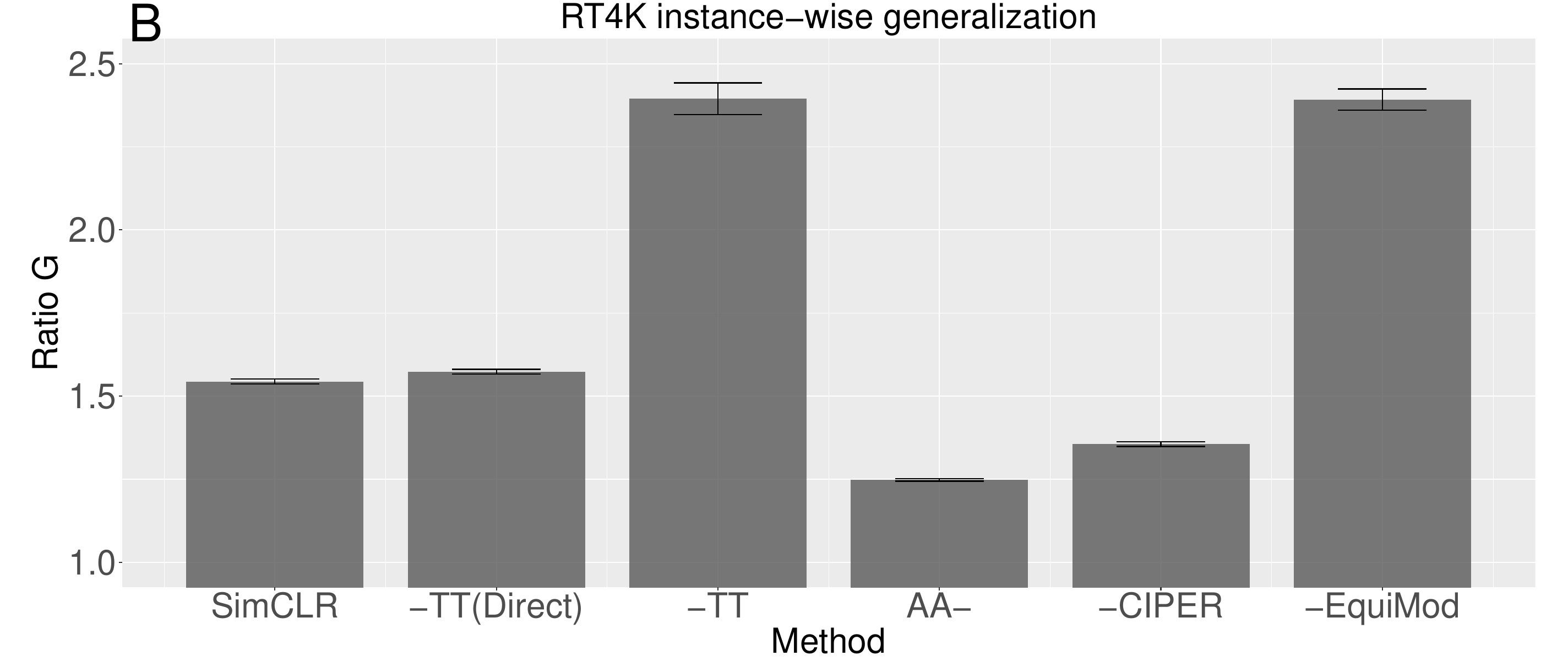}
    \caption{A) Illustration of how we compute our object-wise versus view-wise category generalization metric for a given view of a chair. B) Level of viewpoint invariance versus view-wise category generalization on RT4K after training with methods based on SimCLR.}
    \label{fig:ratio}
\end{figure*}

To quantitavely compare the amount of view-wise category generalization versus viewpoint invariance, we compute the averaged ratio 
\begin{equation}
G(o) = \frac{1}{|C||V|} \sum_{\substack{c \in C \\ v \in V}} \frac{1}{|O_c|}\sum_{o \in O_c} \frac{cos(h_v^o, h_{v+90}^o)}{\sum_{o' \in O_c} cos(h_v^o, h_v^{o'})},
\end{equation}
where $C$, $V$ and $O_c$ are respectively the sets of categories, views and objects in category $c$. The lower $G$, the closer a view's representation is to the same view of different objects from same category versus a +90° rotated view of the very same same object. \figref{fig:ratio}A) illustrates how we compute the contribution to $G$ of a particular view of an object. \figref{fig:ratio}B) shows that AA-SimCLR demonstrates a better ratio than competing methods, thereby validating our qualitative analysis.

\subsection{AA-SSL is more robust to missing data-augmentations}

\begin{figure}
    \centering
    \includegraphics[width=1\linewidth]{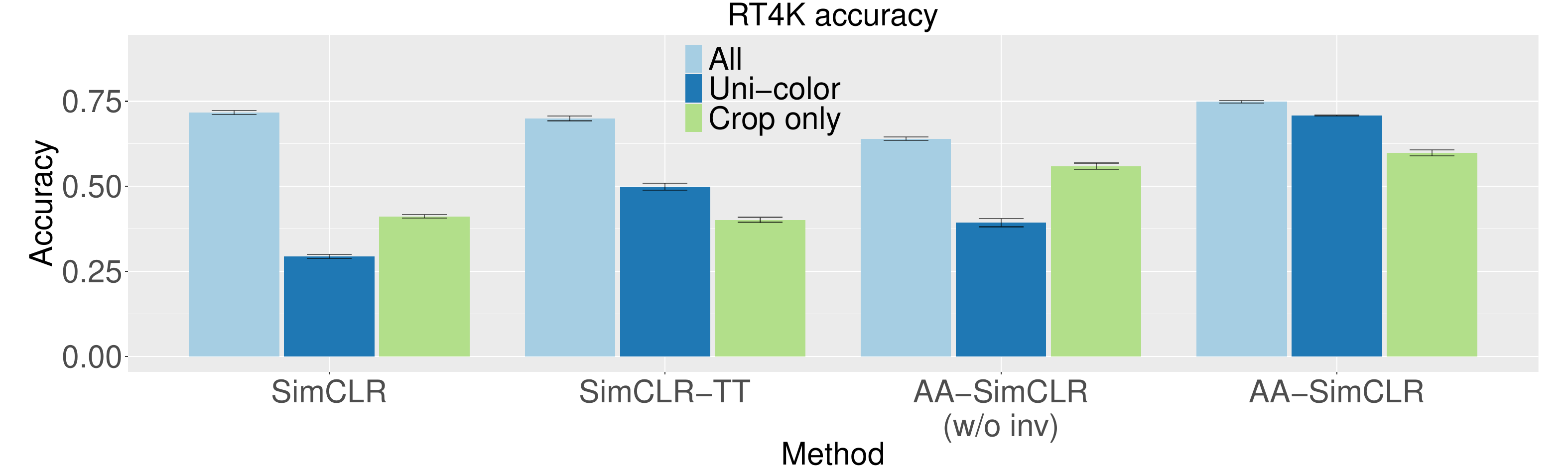}
    \caption{RT4K category accuracy when using reduced sets of data-augmentations during training. AA-SimCLR shows the highest robustness when one or more data-augmentations are removed.}
    \label{fig:robustness}
\end{figure}

Invariance-based methods tend to shortcut the learning process by extracting spurious features from images (texture or color information), which hinders category learning \cite{geirhos2020shortcut,chen2020simple}. For instance, such spurious features may cause the apparent over-fitting discovered in \secref{sec:invariance}. Here, we conjecture that extracting features that are informative about the executed action may mitigate this effect. In \figureautorefname~\ref{fig:robustness}, we reproduce the RT4K experiments of SimCLR, SimCLR-TT and AA-SimCLR, but keeping only the crop/resize augmentation (Crop) or applying the same colors transformations between two positive pairs (Uni-color). We also add a version of AA-SimCLR that only predicts the action representation without invariance learning, called ``AA-SimCLR(w/o inv)''. While the accuracy of SimCLR-TT drops substantially, AA-SimCLR and AA-SimCLR(w/o inv) still support robust categorization. This further highlights the benefits of action-aware representation learning relative to pure invariance learning.

\section{Conclusion}

The learning process of most of today's computer vision systems differs strikingly from that of biological organisms including humans. While the former passively observe large datasets presented to them in order to extract patterns in these data, the latter actively engage with the world and learn to see while interacting with it. This difference could be important, as it allows biological vision systems to actively probe the causal structure of the world rather than merely passively observe correlations. Here we have investigated what benefits may arise from interacting with the world during self-supervised visual representation learning. To this end, we have proposed a new method for action-aware self-supervised learning (AA-SSL) that learns to align representations of actions of an observer interacting with objects with the visual consequences of these actions. This permits the action-related information to influence the latent visual representation, going beyond previous approaches that aim to predict next frames in videos such as PredNet \cite{lotter2016deep}. It also differs from previous work that tries to infer the executed action (or image augmentations) from the images before and after the action \cite{ciperxu}.

We have compared AA-SSL to several SOTA SSL methods on five datasets that provide actions (RT4K, CO3D, MVImgNet-F, MVImgNet-S, MVImgNet-T). Our results show that AA-SSL tends to improve generalization over categories, in particular with limited data. Our analyses reveal that AA-SSL obtains a trade-off between viewpoint invariance vs.\ viewpoint sensitivity that is beneficial for categorization. Furthermore, AA-SSL facilitates the alignment of representations of similar views of different objects from the same category (\figureautorefname~\ref{fig:pacmap}). We also demonstrated that AA-SSL is more robust to the removal of data augmentations and the architecture of the projection heads.



Our work suggests that interactions with objects may partly explain humans' remarkable ability to learn rapidly from few interactions with objects of a particular category. But additional factors beyond what we have considered here may also be crucial. First, humans also rapidly learn to segment objects from the background during such interactions. Second, they actively choose how to interact with objects about which they are learning. For instance, research has shown that humans tend to rotate objects towards planar views \cite{pereira2010early}, which may aid their learning process. This suggests exciting avenues for future work  further advancing unsupervised visual representation learning.






\section{Acknowledgement}

This work was funded by the Deutsche Forschungsgemeinschaft (DFG project ``Abstract REpresentations in Neural Architectures (ARENA)''), as well as the projects ``The Adaptive Mind'' and ``The Third Wave of Artificial Intelligence'' funded by the Excellence Program of the Hessian Ministry of Higher Education, Science, Research and Art (HMWK). JT was supported by the Johanna Quandt foundation. We gratefully acknowledge support from GENCI–IDRIS (Grant 2023-AD011014008), the Mésocentre Clermont-Auvergne of the Université Clermont Auvergne and the Goethe-University (NHR Center NHR@SW) for providing computing and data-processing resources needed for this work.

%
%
\bibliographystyle{splncs04}
\bibliography{references}

\begin{thebibliography}{10}
\providecommand{\url}[1]{\texttt{#1}}
\providecommand{\urlprefix}{URL }
\providecommand{\doi}[1]{https://doi.org/#1}

\bibitem{agrawal2015learning}
Agrawal, P., Carreira, J., Malik, J.: Learning to see by moving. In: Proceedings of the IEEE international conference on computer vision. pp. 37--45 (2015)

\bibitem{agrawal2016learning}
Agrawal, P., Nair, A.V., Abbeel, P., Malik, J., Levine, S.: Learning to poke by poking: Experiential learning of intuitive physics. Advances in neural information processing systems  \textbf{29} (2016)

\bibitem{aubret2022time}
Aubret, A., Ernst, M.R., Teuli{\`e}re, C., Triesch, J.: Time to augment self-supervised visual representation learning. In: The Eleventh International Conference on Learning Representations (2022)

\bibitem{bardes2022vicreg}
Bardes, A., Ponce, J., Lecun, Y.: {VICR}eg: Variance-invariance-covariance regularization for self-supervised learning. In: Proceedings of the 10th International Conference on Learning Representations (ICLR) (2022)

\bibitem{bhardwaj2022steerable}
Bhardwaj, S., McClinton, W., Wang, T., Lajoie, G., Sun, C., Isola, P., Krishnan, D.: Steerable equivariant representation learning  (2022)

\bibitem{chen2020simple}
Chen, T., Kornblith, S., Norouzi, M., Hinton, G.: A simple framework for contrastive learning of visual representations. In: III, H.D., Singh, A. (eds.) Proceedings of the 37th International Conference on Machine Learning. Proceedings of Machine Learning Research, vol.~119, pp. 1597--1607. PMLR (13--18 Jul 2020)

\bibitem{chen2020big}
Chen, T., Kornblith, S., Swersky, K., Norouzi, M., Hinton, G.E.: Big self-supervised models are strong semi-supervised learners. Advances in neural information processing systems  \textbf{33},  22243--22255 (2020)

\bibitem{dangovski2021equivariant}
Dangovski, R., Jing, L., Loh, C., Han, S., Srivastava, A., Cheung, B., Agrawal, P., Solja{\v{c}}i{\'c}, M.: Equivariant contrastive learning. arXiv preprint arXiv:2111.00899  (2021)

\bibitem{devillers2022equimod}
Devillers, A., Lefort, M.: Equimod: An equivariance module to improve visual instance discrimination. In: The Eleventh International Conference on Learning Representations (2022)

\bibitem{doersch2017multi}
Doersch, C., Zisserman, A.: Multi-task self-supervised visual learning. In: Proceedings of the IEEE international conference on computer vision. pp. 2051--2060 (2017)

\bibitem{dong2022viewfool}
Dong, Y., Ruan, S., Su, H., Kang, C., Wei, X., Zhu, J.: Viewfool: Evaluating the robustness of visual recognition to adversarial viewpoints. Advances in Neural Information Processing Systems  \textbf{35},  36789--36803 (2022)

\bibitem{efroni2021provably}
Efroni, Y., Misra, D., Krishnamurthy, A., Agarwal, A., Langford, J.: Provably filtering exogenous distractors using multistep inverse dynamics. In: International Conference on Learning Representations (2021)

\bibitem{franzius2011invariant}
Franzius, M., Wilbert, N., Wiskott, L.: Invariant object recognition and pose estimation with slow feature analysis. Neural computation  \textbf{23}(9),  2289--2323 (2011)

\bibitem{gan2021threedworld}
Gan, C., Schwartz, J., Alter, S., Mrowca, D., Schrimpf, M., Traer, J., De~Freitas, J., Kubilius, J., Bhandwaldar, A., Haber, N., et~al.: Threedworld: A platform for interactive multi-modal physical simulation. In: Thirty-fifth Conference on Neural Information Processing Systems Datasets and Benchmarks Track (Round 1) (2021)

\bibitem{garrido2023self}
Garrido, Q., Najman, L., LeCun, Y.: Self-supervised learning of split invariant equivariant representations  (2023)

\bibitem{geirhos2020shortcut}
Geirhos, R., Jacobsen, J.H., Michaelis, C., Zemel, R., Brendel, W., Bethge, M., Wichmann, F.A.: Shortcut learning in deep neural networks. Nature Machine Intelligence  \textbf{2}(11),  665--673 (2020)

\bibitem{gordon2020watching}
Gordon, D., Ehsani, K., Fox, D., Farhadi, A.: Watching the world go by: Representation learning from unlabeled videos. arXiv preprint arXiv:2003.07990  (2020)

\bibitem{grill2020bootstrap}
Grill, J.B., Strub, F., Altch\'{e}, F., Tallec, C., Richemond, P., Buchatskaya, E., Doersch, C., Avila~Pires, B., Guo, Z., Gheshlaghi~Azar, M., Piot, B., kavukcuoglu, k., Munos, R., Valko, M.: Bootstrap your own latent - a new approach to self-supervised learning. In: Larochelle, H., Ranzato, M., Hadsell, R., Balcan, M., Lin, H. (eds.) Advances in Neural Information Processing Systems. vol.~33, pp. 21271--21284. Curran Associates, Inc. (2020), \url{https://proceedings.neurips.cc/paper/2020/file/f3ada80d5c4ee70142b17b8192b2958e-Paper.pdf}

\bibitem{gupta2023learning}
Gupta, S., Robinson, J., Lim, D., Villar, S., Jegelka, S.: Learning structured representations with equivariant contrastive learning  (2023)

\bibitem{harman1999active}
Harman, K.L., Humphrey, G.K., Goodale, M.A.: Active manual control of object views facilitates visual recognition. Current Biology  \textbf{9}(22),  1315--1318 (1999)

\bibitem{hayward2003after}
Hayward, W.G.: After the viewpoint debate: where next in object recognition? Trends in cognitive sciences  \textbf{7}(10),  425--427 (2003)

\bibitem{he2020momentum}
He, K., Fan, H., Wu, Y., Xie, S., Girshick, R.: Momentum contrast for unsupervised visual representation learning. In: Proceedings of the IEEE/CVF Conference on Computer Vision and Pattern Recognition (CVPR) (June 2020)

\bibitem{hernandez2023visual}
Hernandez, J., Villegas, R., Ordonez, V.: Visual representation learning from unlabeled video using contrastive masked autoencoders. arXiv preprint arXiv:2303.12001  (2023)

\bibitem{jayaraman2015learning}
Jayaraman, D., Grauman, K.: Learning image representations tied to ego-motion. In: Proceedings of the IEEE International Conference on Computer Vision. pp. 1413--1421 (2015)

\bibitem{jayaraman2016look}
Jayaraman, D., Grauman, K.: Look-ahead before you leap: end-to-end active recognition by forecasting the effect of motion. In: Computer Vision--ECCV 2016: 14th European Conference, Amsterdam, The Netherlands, October 11-14, 2016, Proceedings, Part V 14. pp. 489--505. Springer (2016)

\bibitem{jayaraman2016slow}
Jayaraman, D., Grauman, K.: Slow and steady feature analysis: higher order temporal coherence in video. In: Proceedings of the IEEE Conference on Computer Vision and Pattern Recognition. pp. 3852--3861 (2016)

\bibitem{jayaraman2017learning}
Jayaraman, D., Grauman, K.: Learning image representations tied to egomotion from unlabeled video. International Journal of Computer Vision  \textbf{125},  136--161 (2017)

\bibitem{jenni2021time}
Jenni, S., Jin, H.: Time-equivariant contrastive video representation learning. In: Proceedings of the IEEE/CVF International Conference on Computer Vision. pp. 9970--9980 (2021)

\bibitem{kim2019emi}
Kim, H., Kim, J., Jeong, Y., Levine, S., Song, H.O.: Emi: Exploration with mutual information. In: International Conference on Machine Learning. pp. 3360--3369. PMLR (2019)

\bibitem{knights2021temporally}
Knights, J., Harwood, B., Ward, D., Vanderkop, A., Mackenzie-Ross, O., Moghadam, P.: Temporally coherent embeddings for self-supervised video representation learning. In: 2020 25th International Conference on Pattern Recognition (ICPR). pp. 8914--8921. IEEE (2021)

\bibitem{landau1988importance}
Landau, B., Smith, L.B., Jones, S.S.: The importance of shape in early lexical learning. Cognitive development  \textbf{3}(3),  299--321 (1988)

\bibitem{lee2021improving}
Lee, H., Lee, K., Lee, K., Lee, H., Shin, J.: Improving transferability of representations via augmentation-aware self-supervision. Advances in Neural Information Processing Systems  \textbf{34},  17710--17722 (2021)

\bibitem{liang2023alp}
Liang, X., Han, A., Yan, W., Raghunathan, A., Abbeel, P.: Alp: Action-aware embodied learning for perception. arXiv preprint arXiv:2306.10190  (2023)

\bibitem{lotter2016deep}
Lotter, W., Kreiman, G., Cox, D.: Deep predictive coding networks for video prediction and unsupervised learning. arXiv preprint arXiv:1605.08104  (2016)

\bibitem{mittal2022learning}
Mittal, H., Morgado, P., Jain, U., Gupta, A.: Learning state-aware visual representations from audible interactions. Advances in Neural Information Processing Systems  \textbf{35},  23765--23779 (2022)

\bibitem{morgado2021audio}
Morgado, P., Vasconcelos, N., Misra, I.: Audio-visual instance discrimination with cross-modal agreement. In: Proceedings of the IEEE/CVF Conference on Computer Vision and Pattern Recognition. pp. 12475--12486 (2021)

\bibitem{orhan2020self}
Orhan, E., Gupta, V., Lake, B.M.: Self-supervised learning through the eyes of a child. In: Larochelle, H., Ranzato, M., Hadsell, R., Balcan, M., Lin, H. (eds.) Advances in Neural Information Processing Systems. vol.~33, pp. 9960--9971. Curran Associates, Inc. (2020), \url{https://proceedings.neurips.cc/paper/2020/file/7183145a2a3e0ce2b68cd3735186b1d5-Paper.pdf}

\bibitem{parthasarathy2022self}
Parthasarathy, N., Eslami, S., Carreira, J., H{\'e}naff, O.J.: Self-supervised video pretraining yields strong image representations. arXiv preprint arXiv:2210.06433  (2022)

\bibitem{parthasarathy2023self}
Parthasarathy, N., Eslami, S.A., Carreira, J., Henaff, O.J.: Self-supervised video pretraining yields robust and more human-aligned visual representations. In: Thirty-seventh Conference on Neural Information Processing Systems (2023)

\bibitem{pathak2017curiosity}
Pathak, D., Agrawal, P., Efros, A.A., Darrell, T.: Curiosity-driven exploration by self-supervised prediction. In: International Conference on Machine Learning (ICML). vol.~2017 (2017)

\bibitem{pereira2010early}
Pereira, A.F., James, K.H., Jones, S.S., Smith, L.B.: Early biases and developmental changes in self-generated object views. Journal of vision  \textbf{10}(11),  22--22 (2010)

\bibitem{przewikezlikowski2023augmentation}
Przewi{{e}}{\'z}likowski, M., Pyla, M., Zieli{\'n}ski, B., Twardowski, B., Tabor, J., {\'S}mieja, M.: Augmentation-aware self-supervised learning with guided projector. arXiv preprint arXiv:2306.06082  (2023)

\bibitem{purushwalkam2020demystifying}
Purushwalkam, S., Gupta, A.: Demystifying contrastive self-supervised learning: Invariances, augmentations and dataset biases. Advances in Neural Information Processing Systems  \textbf{33},  3407--3418 (2020)

\bibitem{radford2021learning}
Radford, A., Kim, J.W., Hallacy, C., Ramesh, A., Goh, G., Agarwal, S., Sastry, G., Askell, A., Mishkin, P., Clark, J., et~al.: Learning transferable visual models from natural language supervision. In: International conference on machine learning. pp. 8748--8763. PMLR (2021)

\bibitem{reizenstein2021common}
Reizenstein, J., Shapovalov, R., Henzler, P., Sbordone, L., Labatut, P., Novotny, D.: Common objects in 3d: Large-scale learning and evaluation of real-life 3d category reconstruction. In: Proceedings of the IEEE/CVF International Conference on Computer Vision. pp. 10901--10911 (2021)

\bibitem{sanyal2023computational}
Sanyal, D., Michelson, J., Yang, Y., Ainooson, J., Kunda, M.: A computational account of self-supervised visual learning from egocentric object play. arXiv preprint arXiv:2305.19445  (2023)

\bibitem{schneider2021contrastive}
Schneider, F., Xu, X., Ernst, M.R., Yu, Z., Triesch, J.: Contrastive learning through time. In: SVRHM 2021 Workshop @ NeurIPS (2021)

\bibitem{stojanov2021using}
Stojanov, S., Thai, A., Rehg, J.M.: Using shape to categorize: Low-shot learning with an explicit shape bias. In: Proceedings of the IEEE/CVF Conference on Computer Vision and Pattern Recognition (CVPR). pp. 1798--1808 (June 2021)

\bibitem{tan2024egodistill}
Tan, S., Nagarajan, T., Grauman, K.: Egodistill: Egocentric head motion distillation for efficient video understanding. Advances in Neural Information Processing Systems  \textbf{36} (2024)

\bibitem{tarr1998three}
Tarr, M.J., Williams, P., Hayward, W.G., Gauthier, I.: Three-dimensional object recognition is viewpoint dependent. Nature neuroscience  \textbf{1}(4),  275--277 (1998)

\bibitem{tschannen2020self}
Tschannen, M., Djolonga, J., Ritter, M., Mahendran, A., Houlsby, N., Gelly, S., Lucic, M.: Self-supervised learning of video-induced visual invariances. In: Proceedings of the IEEE/CVF Conference on Computer Vision and Pattern Recognition. pp. 13806--13815 (2020)

\bibitem{vuilleumier2002multiple}
Vuilleumier, P., Henson, R., Driver, J., Dolan, R.J.: Multiple levels of visual object constancy revealed by event-related fmri of repetition priming. Nature neuroscience  \textbf{5}(5),  491--499 (2002)

\bibitem{wang2023distortion}
Wang, J., Song, S., Su, J., Zhou, S.K.: Distortion-disentangled contrastive learning. arXiv preprint arXiv:2303.05066  (2023)

\bibitem{wang2016actions}
Wang, X., Farhadi, A., Gupta, A.: Actions\~{} transformations. In: Proceedings of the IEEE conference on Computer Vision and Pattern Recognition. pp. 2658--2667 (2016)

\bibitem{wang2021understanding}
Wang, Y., Huang, H., Rudin, C., Shaposhnik, Y.: Understanding how dimension reduction tools work: an empirical approach to deciphering t-sne, umap, trimap, and pacmap for data visualization. The Journal of Machine Learning Research  \textbf{22}(1),  9129--9201 (2021)

\bibitem{wiskott2002slow}
Wiskott, L., Sejnowski, T.J.: Slow feature analysis: Unsupervised learning of invariances. Neural Computation  \textbf{14}(4),  715--770 (2002). \doi{10.1162/089976602317318938}

\bibitem{wu2021contrastive}
Wu, H., Wang, X.: Contrastive learning of image representations with cross-video cycle-consistency. In: Proceedings of the IEEE/CVF International Conference on Computer Vision. pp. 10149--10159 (2021)

\bibitem{xiao2020noise}
Xiao, K.Y., Engstrom, L., Ilyas, A., Madry, A.: Noise or signal: The role of image backgrounds in object recognition. In: International Conference on Learning Representations (2020)

\bibitem{xie2022should}
Xie, Y., Wen, J., Lau, K.W., Rehman, Y.A.U., Shen, J.: What should be equivariant in self-supervised learning. In: Proceedings of the IEEE/CVF Conference on Computer Vision and Pattern Recognition. pp. 4111--4120 (2022)

\bibitem{xu2021rethinking}
Xu, J., Wang, X.: Rethinking self-supervised correspondence learning: A video frame-level similarity perspective. In: Proceedings of the IEEE/CVF International Conference on Computer Vision (ICCV). pp. 10075--10085 (October 2021)

\bibitem{ciperxu}
Xu, X., Triesch, J.: Ciper: Combining invariant and equivariant representations using contrastive and predictive learning. In The 32nd International Conference on Artificial Neural Networks  (2023)

\bibitem{yu2023mvimgnet}
Yu, X., Xu, M., Zhang, Y., Liu, H., Ye, C., Wu, Y., Yan, Z., Zhu, C., Xiong, Z., Liang, T., et~al.: Mvimgnet: A large-scale dataset of multi-view images. In: Proceedings of the IEEE/CVF Conference on Computer Vision and Pattern Recognition. pp. 9150--9161 (2023)

\bibitem{zambelli2021learning}
Zambelli, M., Aytar, Y., Visin, F., Zhou, Y., Hadsell, R.: Learning rich touch representations through cross-modal self-supervision. In: Conference on Robot Learning. pp. 1415--1425. PMLR (2021)

\end{thebibliography}




\newpage

\section{Supplementary analysis}

\subsection{Comparison of AA-SimCLR with supervised learning}

\begin{table}
    \centering
    \begin{tabular}{c|c|c|c|c}
         & RT4K & MV-T & MV-S & MV-F \\
         \midrule
        \textbf{AA-} & $74.8 \pm 0.3$ & $42.8 \pm 0.8$ & $78.4 \pm 0.1$ & $96.1$ \\
         Sup & $73.6 \pm 0.3$& $38.3 \pm 0.9$& $78.8 \pm 0.25$& $98$
    \end{tabular}
    \caption{Comparison of AA-SimCLR to supervised baseline.}
    \label{tab:supervised}
\end{table}

In \tableautorefname~\ref{tab:supervised}, we compare our method with a supervised baseline that use the same augmentations. We observe a small performance gap between AA-SimCLR and the supervised oracle. Interestingly, our method outperforms the supervised baseline on small datasets.

\subsection{Lambda hyper-parameter does not require careful tuning}

\begin{figure}
    \centering
    \includegraphics[width=0.7\linewidth]{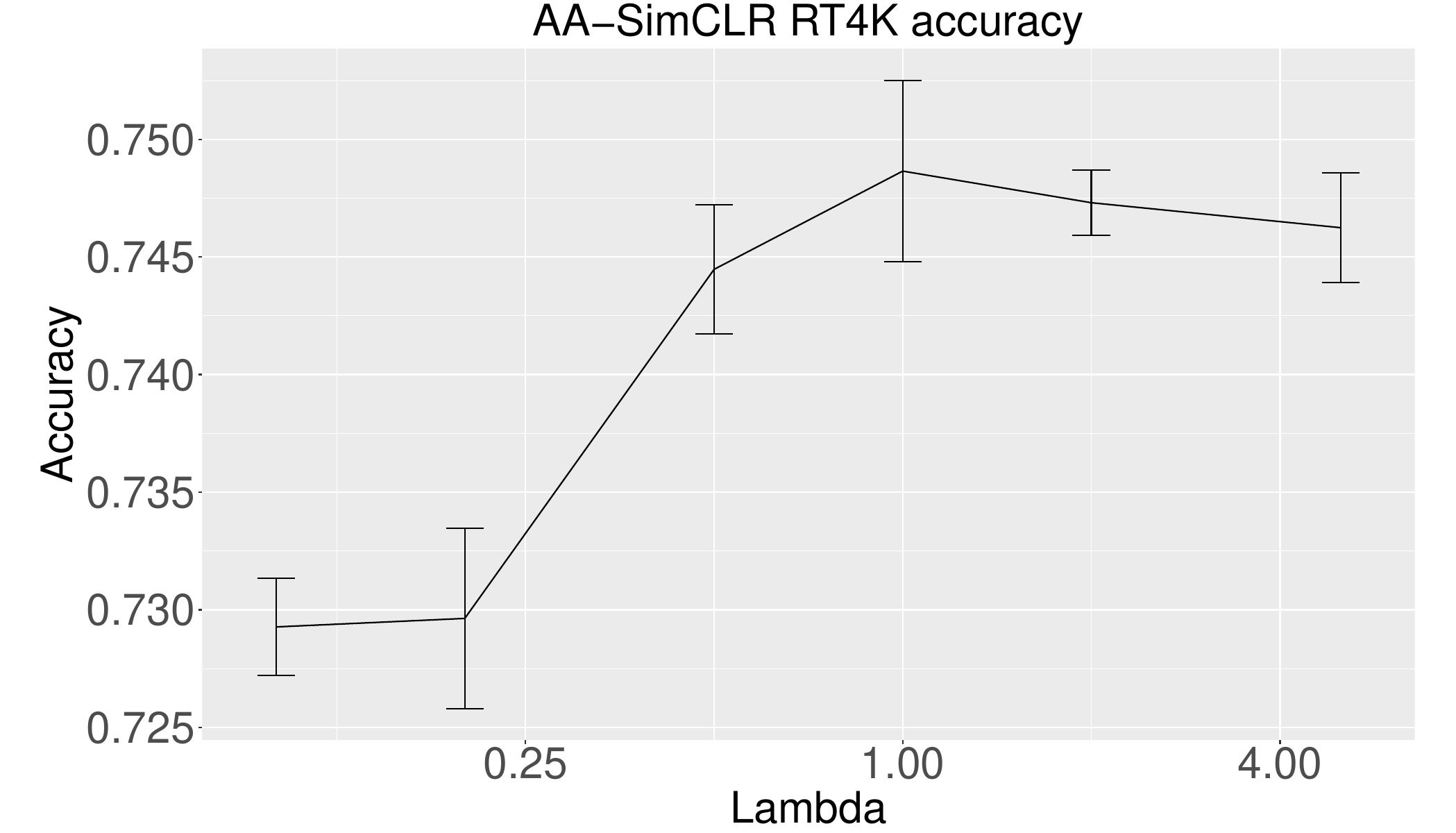}
    \caption{Impact of the hyper-parameter $\lambda$ on the test category accuracy of AA-SimCLR on RT4K.}
    \label{fig:lambda}
\end{figure}

The hyper-parameter $ \lambda $ of AA-SSL balances the action-aware part of the loss function against the invariant part. In our main results, we did not fine tune $ \lambda $ for each dataset. Here, we study the effect of varying $ \lambda $ for the RT4K dataset. In \figref{fig:lambda}, we observe an asymmetric inverted U-shape of the classification accuracy as a function of $ \lambda $, with the maximum being obtained at $ \lambda = 1 $. However, the quality of the representation seems robust against small changes of $ \lambda $, in particular when considering larger values of $ \lambda $. Thus, we do not advocate for a careful tuning and recommend using $\lambda=1$.

\subsection{PaCMAP visualization for different categories}\label{app:morepacmaps}

\begin{figure*}
    \centering
    \includegraphics[width=1\linewidth]{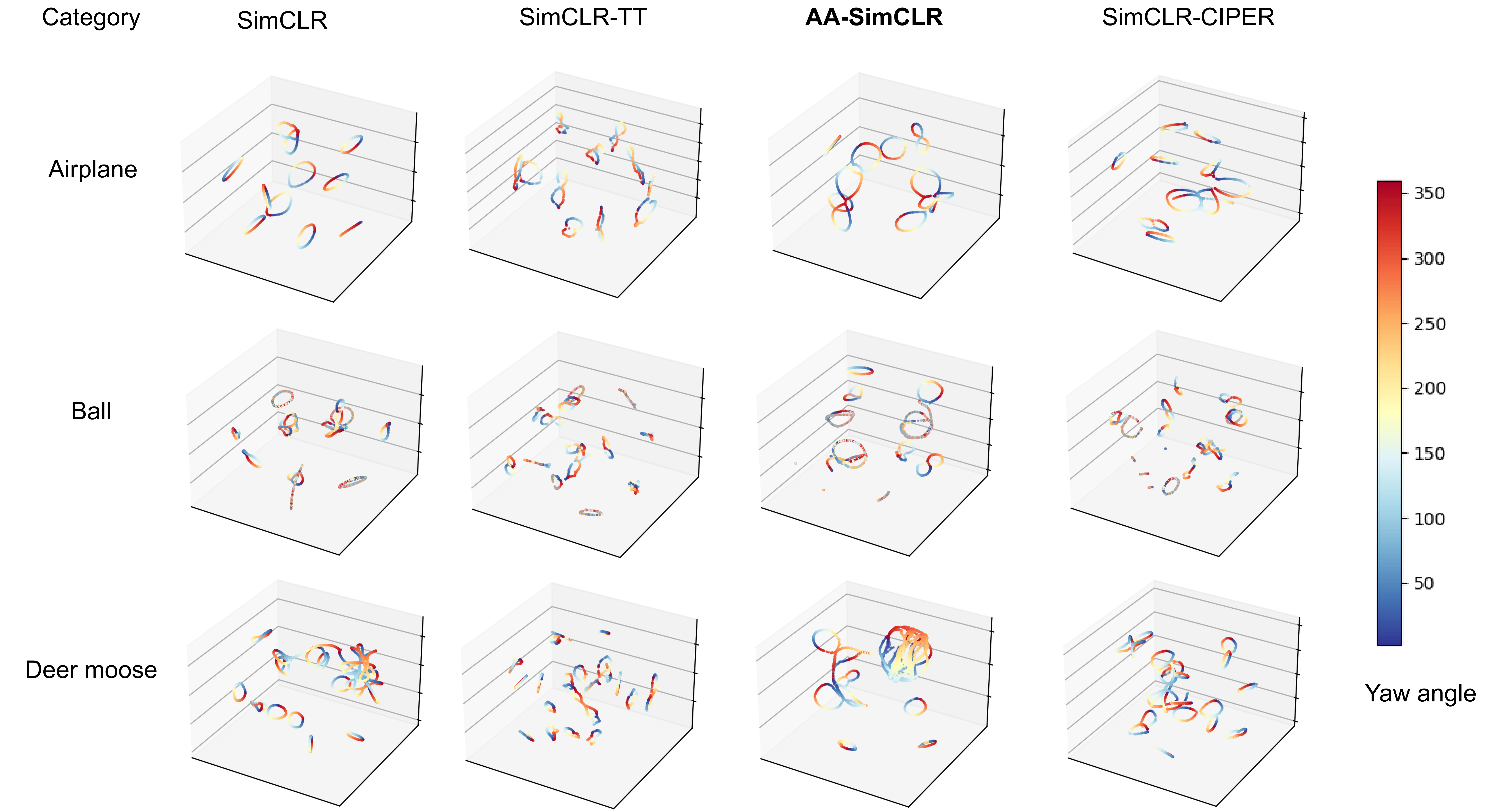}
    \caption{View-wise PaCMAP visualization of learnt representations of RT4K test images extracted from 3 categories. Categories were selected to reflect different properties of categories. Airplanes share the same structural properties (two perpendicular axes) but highly variable features (single wing versus double wings); balls are often rotationally symmetric, making it hard to distinguish viewpoints based on an object's shape; deers have a similar global shape and inter-object differences are small, e.g., the exact shapes of the antlers may differ.}
    \label{fig:allpacmap}
\end{figure*}

In \secref{sec:quality}, we showed the PaCMAP visualization for the (most abundant) object category ``chairs.'' In \figref{fig:allpacmap}, we plot the same view-wise visualization for three other categories. We observe that, for some categories like ``ball'' or ``airplane'', none of the methods manage to generalize based on viewpoints. For the ``deer moose'' category, we obtain similar results as for the ``chair'' category in \secref{sec:quality}: only AA-SimCLR manages to align representations of different objects viewed in the same orientation. Overall, this suggests there is a margin of improvement to further enforce view-wise category generalization.

\end{document}